# Robust Support Vector Machines for Speaker Verification Task

Kawthar Yasmine ZERGAT[1], Abderrahmane AMROUCHE[1]

[1] `Speech Com. & Signal Proc. Lab.-LCPTS`
`Faculty of Electronics and Computer Sciences,`
`USTHB, Bab Ezzouar, 16 111, Algeria.`

**Abstract**

An important step in speaker verification is extracting features that best characterize the speaker voice. This paper investigates a front-end processing that aims at improving the performance of speaker verification based on the SVMs classifier, in text independent mode. This approach combines features based on conventional Mel-cepstral Coefficients (MFCCs) and Line Spectral Frequencies (LSFs) to constitute robust multivariate feature vectors. To reduce the high dimensionality required for training these feature vectors, we use a dimension reduction method called principal component analysis (PCA). In order to evaluate the robustness of these systems, different noisy environments have been used. The obtained results using TIMIT database showed that, using the paradigm that combines these spectral cues leads to a significant improvement in verification accuracy, especially with PCA reduction for low signal-to-noise ratio noisy environment.

## 1. Introduction

A typical speaker verification system usually consists of two phases: an enrollment phase, and an Authentication phase. In the enrollment phase, the system extracts speaker-specific information from the speech signal to be used to build a model for the speaker [1], where the purpose of the testing phase is to determine whether the speech samples belong to the person that claims his/her identity or not.

In all audio processing, the speech input is converted into a feature vector representation [2]. Linear Prediction Cepstral Coefficients (LPCC) and Perceptual Linear Prediction Cepstral Coefficients (PLPCC), the Mel Frequency Cepstral Coefficients (MFCC) [3] approach has been the most employed for feature extraction. For modeling, Support Vector Machine (SVM) [4] represents a discriminative classifier which has achieved impressive results in several pattern recognition tasks. Indeed, the SVMs are interesting because they discriminate between classes (speaker/impostor) and could be used to train non-linear decision boundaries in an efficient manner. In this paper, for speaker verification, we investigate on SVM classifier based on Principal Component Analysis (PCA) [5] to get the efficiently reduced dimension of feature vectors. First, the MFCC, Line Spectral Frequency (LSF) features are extracted from the speech voice sample. The concatenation of these features vectors (MFCC-LSF) is made. Secondly, the new feature vectors with reduced dimension are obtained by applying PCA dimensionality reduction to each speaker vectors. Finally, theses transformed feature vectors are used as input to the SVM system for text independent speaker verification task. To validate the influence of PCA dimensionality reduction, we have evaluated the robustness of both SVM and PCA-SVM systems in different noisy environments at different levels of SNRs. The rest of the paper is as follows. In sections 2, we describe the Feature Extraction process used and discuss the principles of SVM in section 3. Section 4 and 5 are the experimental setup and the results of the experiments conducted on a subset of TIMIT Database. Finally, we conclude in Section 6.

## 2. Feature Extraction

2.1 Mel Frequency Cepstral Coefficient (MFCC)

MFCCs were introduced in early 1980s for speech recognition applications. The key steps involved in computing MFCC features are shown in Fig. 1. The speech signal is first pre-emphasized by applying the following filter [1],

$$x(t) = y(t) - ay(t-1), \text{Where } a \in [0.95, 0.98] \quad (1)$$

The goal of the filter is to enhance the high frequencies of the spectrum, which is diminished during the speech production process. Following the pre-emphasis stage is a windowing step, the speech samples are weighed by a suitable windowing function, The Hamming window is extensively used in speaker verification to taper the original signal on the sides witch reduce the side effects [6].

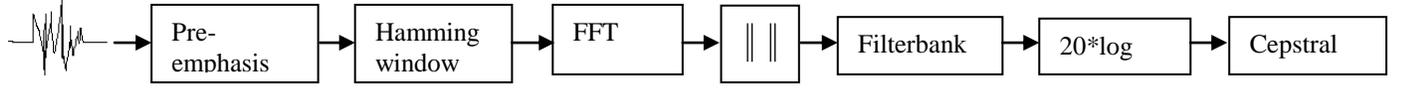

Fig. 1 MFCC features extraction.

The result of windowing the signal is shown below:

$$W(n) = 0.54 - 0.46\cos\left[\frac{2\Pi n}{N-1}\right], \quad 0 \leq n \leq N-1 \quad (2)$$

Once the speech signal has been windowed its fast Fourier transform (FFT) is calculated. Finally, the modulus of the FFT is extracted and a power spectrum is obtained [6]. The obtained spectrum is then multiplied by filterbank.
The filters that are generally used in MFCC computation are triangular filters, and their center frequencies are chosen according a logarithmic frequency scale also known as Mel-frequency scale witch conforms to response observed in human auditory systems. The localization of the central frequencies of the filters is given by [1]:

$$f_{MEL} = 1000 \cdot \frac{\log(1 + f_{LIN}/1000)}{\log 2} \quad (3)$$

An additional transform, is to obtain the spectral vectors by taking the log of the spectral envelope and multiply each coefficient by 20 in order to get the spectral envelope in dB [6]. Finally, the cosine discrete transform (DCT) is applied to the spectral vectors witch yields cepstral coefficients frequencies, and is given by [6]:

$$c_n = \sum_{k=1}^{K} S_k \cos[n(k-\frac{1}{2})\frac{\pi}{k}], \quad n=1,2,....,L \quad (4)$$

Where $K$ is the number of log-spectral coefficients calculated in previous step, $S_k$ are the log-spectral coefficients, and $L$ is the number of cepstral coefficients to calculate.

2.2 Line Spectral Frequency Cues (LSF)

The starting point for deriving the LSF's is the response of the prediction error filter.

$$A(z) = 1 - \sum_{k=1}^{P} a_k z^{-k} \quad (5)$$

Where $P$ represents the prediction order, and $a_k$ are the LPC filter coefficients. In the LPC the mean squared error between the linearly predicted speech sample and the actual one is minimized over a finite interval [7]. The transfer function of the LPC filter with a gain $G$ is given by:

$$H(z) = \frac{G}{1 + \sum_{k=1}^{P} a_k z^{-k}}, \quad (6)$$

From $H(z)$, a symmetric polynomial $S_{P+1}(z)$ and an antisymmetric polynomial $\theta_{P+1}(z)$ are calculated by adding and subtracting the time-reversed system function.

$$S_{P+1} = A_p(z) + z^{-(p+1)} A_p(z^{-1}), \text{ And}$$
$$\theta_{p+1} = A_p(z) - z^{-(p+1)} A_p(z^{-1}), \quad (7)$$

The polynomials contain trivial zeros for even values of $p$ at $z = -1$ and at $z = 1$. These roots can be removed in order to derive the following quantities [7]:

$$\tilde{S}(z) = \frac{S_{P+1}(z)}{(1+z)} = \omega_0 z^p + \omega_1 z^{p-1} + ... + \omega_p, \quad \text{And}$$
$$\tilde{\theta}(z) = \frac{\theta_{P+1}(z)}{(1-z)} = \sigma_0 z^p + \sigma_1 z^{p-1} + ... + \sigma_p, \quad (8)$$

The LSFs are the roots of $\tilde{S}(z)$ and $\tilde{\theta}(z)$ and alternate with each other on the unit circle [7].

## 3. Support Vector Machine

Support Vector Machine (SVM) is a binary linear classifier in its basic form. It has been recently adopted in speaker recognition task. Given a set of linearly separable two-class training data, there are many possible solutions for a discriminative classifier [8]. An SVM seeks to find the Optimal Separating Hyperplane (OSH) between classes by focusing on the training cases that lie at the edge of the class distributions, the support vectors, with the other training cases effectively discarded [8]. Formally, the discriminant function of SVM is given by:

$$f(x) = class(x) = sign\left[\sum_{t=i}^{N} \alpha_i y_i K(x, x_i) + b\right] \quad (9)$$

Here $y_i \in \{-1, +1\}$ are the ideal output values,

$\sum_{t=i}^{N} \alpha_i y_i = 0$ and $\alpha_i \geq 0$. The support vectors $x_i$, their

corresponding weights $\alpha_i$ and the bias term b.

To calculate the classification function class (x) we use the dot product in feature space that can also be expressed in the input space by the kernel function $K(\cdot, \cdot)$. Among the most widely used cores we find:

RBF kernel: $k(x, y) = e^{-\gamma \|x - x_i\|^2}$

Polynomial kernel: $k(x, y) = (x^T \cdot y + 1)$

Finally, the classification of data is made as follow:

$$x \in \begin{cases} Classe\ 1 \\ classe\ 2 \end{cases} \text{if} \begin{cases} X \succ 0 \\ Otherwise \end{cases}$$

## 4. Experimental Protocol

### 4.1 Description of the Database

The corpus used in this work is issued from the TIMIT database. This database includes time-aligned orthographic, phonetic and word transcriptions as well as a 16-bit, 16 kHz speech file for each utterance and is recorded in ".ADC" format, where each sentence has 3s of length spoken in English language. We have selected a set of 90 speakers, for both training and testing phases, each of whom reads 5 phonetically rich sentences for training task and 3 utterances for testing task. To simulate the impostors, 50 unknown speakers (25 female and 25 male) are used from the same database (TIMIT) and are different from the 90 speakers used previously, with five utterances spoken by each unknown speaker.

### 4.2 Parameterization Phase

In this work, we have included as many speakers' characteristics as possible. So, our first feature space is made with 12 MFCC coefficients plus Energy parameter and the first and second derivatives, which yield 39-dimensional feature vector, extracted from the middle window every 10ms. A voice activity detector is used to eliminate silence and noise frames from the training and testing signals in order to avoid modeling and detecting the environment rather than the speaker. In the second part, 12 LSF coefficients were extracted. Finally, the first feature space (MFCCs +E+ $\Delta + \Delta\Delta$) was combined with 12 LSF coefficients to constitute a multi-dimensional feature set. The dimension of the combined vectors is then equal to 51. Once the feature vectors have been calculated, they can be centered, using Cepstral Mean Subtraction (CMS), this is carried out by estimating a mean vector for the extracted set of cepstral features and subtracting it from all the feature vectors. In the other hand, the size of the vectors of parameters is an important problem that arises when adding parameters. To address this, technique to reduce the number of parameters was used, these include PCA dimensionality reduction.

### 4.3 Modeling Phase

For the enrolment phase, we did however make use of a RBF kernel function for the SVM classifier. In order to evaluate the performance of the system, two types of additive noise produced by a Speech Babble and a Subway noises reaching high levels of SNR and derived from the NOISEX-92 database (NATO: AC 243/RSG 10) are added to the test speech signal of the TIMIT database. In speaker verification task, there are two types of errors; *false acceptance* (FA) and *false rejection* (FR). The Equal Error Rate (ERR) is the point where the rate of FR's is equal to the rate of FA's. For classification, The Detection Error Tradeoff (DET) curve is a popular way of graphically representing the performance of speaker verification system.

## 5. Experiment Results

### 5.1 Speaker Verification using Original Speech Waveforms

To show the effectiveness of the proposed method, we performed two experiments using speech data without and using the PCA dimensionality reduction for the SVM classifier in speaker verification task. In both experiments we used Equal Error Rate (EER) as the performance criterion. For the first experiment a comparative study shows the contribution of the concatenation between the MFCC and the LSF features. As shown in Fig. 2, the concatenated feature vector brings the less EER equal to 0.54% against the LSF and MFCC feature vectors with an EER equal to 7.39% and 3.69% respectively.

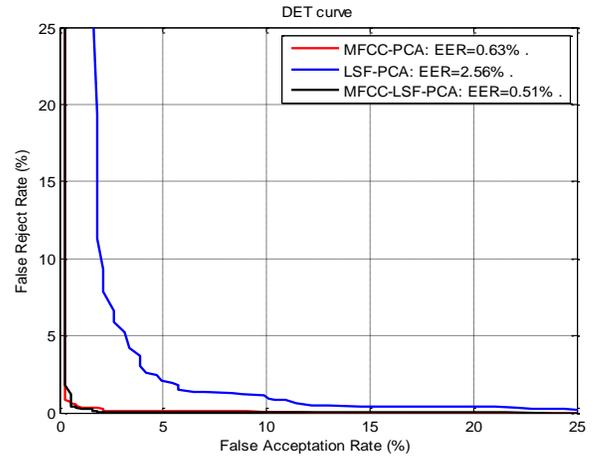

Fig. 3 Performance evaluation for PCA-SVM based speaker verification task.

### 5.2 Verification Accuracy under Noisy Environments

The main goal of the experiments done in this section is the study of the verification performances of both SVM and PCA-SVM systems in different noisy environments, for this, two noisy environments which are Speech Babble and Subway noises were used. We evaluated the error rate by applying dimensionality reduction by PCA algorithm on the concatenated MFCC-LSF feature vectors. The results are shown in the following figures.

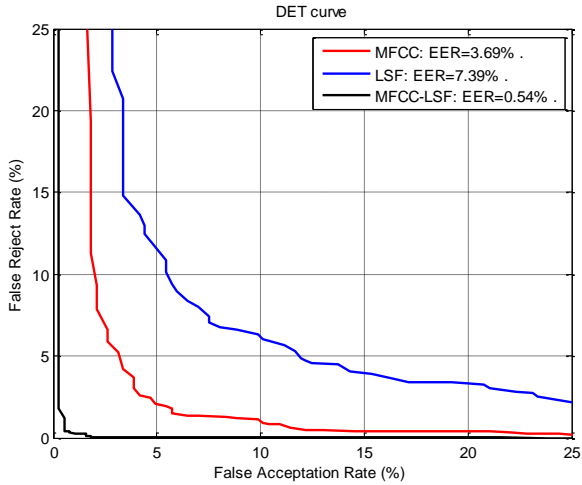

Fig. 2 Performance evaluation for SVM based speaker verification task.

Speaker verification experiment with PCA based SVM classifier has been performed too with these various feature space components. From the following fig. 3, it's clearly seen that PCA improves significantly the recognition accuracy, until an EER=0.51% for the concatenated (MFCC, LSF) feature vectors.

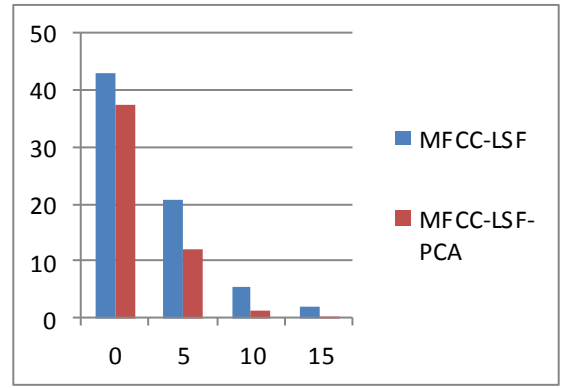

Fig. 4 Performances evaluation for SVM and PCA-SVM in noisy environment corrupted by Babble speech noise.

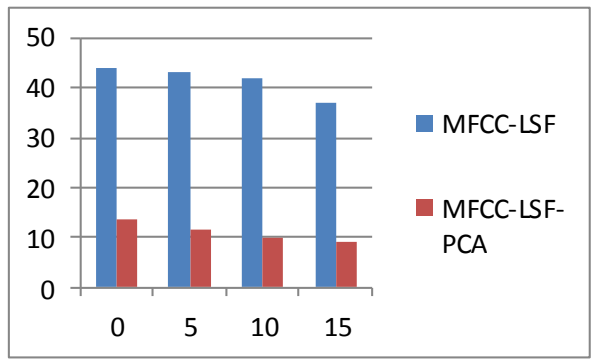

Fig. 5 Performances evaluation for SVM and PCA-SVM in noisy environment corrupted by Subway noise.

As shown in the above figures, it is clearly seen that applying PCA algorithm on feature vectors leads to an interesting increase of speaker verification accuracy. Quantifying the input data by other Algorithms such as LBG and k-means, quantify all the data including the insignificant and repeated items presented in the speech signal [9], by cons, when using PCA dimensionality reduction, we project the data into lower dimensional space, where the low variance components are eliminated. The obtained results confirm the effect of PCA, for example, in case of noisy environment corrupted by Babble speech noise at SNR = 0dB, the EER decrease from 42.87% to 37.39% which represents an interesting improvement in bad conditions.

## 6. Conclusion

This paper has presented and evaluated a text-independent speaker verification systems based on SVM classifier. To attain better performance, two systems were trained, the SVM and the PCA-SVM systems. In this study, we have examined the influence of feature vectors and PCA dimensionality reduction on speaker verification rate in both clean and noisy environments. Carried out on TIMIT database, it is noticed that, the combination between MFCC and LSF outperforms the conventional MFCC parameters. In the other hand, for the PCA-SVM model, the recognition accuracy has increase significantly comparing to the SVM model, especially in hard conditions (SNR= 0dB). We are currently continuing the effort towards the optimization of this system using other dimensionally reduction method.

**Kawthar Yasmine Zergat** Received her Master II degree in Communication and Multimedia from the University of Science and Technology Houari Boumedienne (USTHB), Algiers in 2010. Currently, she is pursuing the Ph.D. degree in, Telecommunications and Information Processing in the Communication Systems and Speech Processing Laboratory, USTHB. Her current research concentrates on robust speaker recognition and speech processing.

**Abderrahmane Amrouche** Was born in Algeria. He received his "diplome d'ingenieur" (engineer degree) in Electronics from the National Polytechnic school of Algiers in 1980. He received his "Magister" degree in 1995 and Doctorat d'Etat" (Ph.D) in Real Time Systems in 2007 from the University of Science and Technology Houari Boumedienne (USTHB). He is an Assistant Professor in Communication Systems and Speech Processing Laboratory, USTHB. His research interests include pattern recognition, speech processing, Multilingual speech recognition, neural networks, prosodic modelling.